\documentclass{article}

\usepackage[final,nonatbib]{neurips_2023}

\usepackage[utf8]{inputenc} 
\usepackage[T1]{fontenc}    
\usepackage{hyperref}       
\usepackage{url}            
\usepackage{booktabs}       
\usepackage{amsfonts}       
\usepackage{amsmath}
\usepackage{nicefrac}       
\usepackage{microtype}      
\usepackage{xcolor}         
\usepackage{wrapfig}
\usepackage{multirow}
\usepackage{graphicx}
\usepackage{graphbox}
\usepackage{subcaption}
\usepackage{algorithm}
\usepackage{algorithmic}

\newcommand{\TODO}[1]{}
\renewcommand{\TODO}[1]{{\color{cyan} [TODO: {#1}]}}
\newcommand{\WIP}[1]{}
\renewcommand{\WIP}[1]{{\color{magenta} [WIP: {#1}]}}
\definecolor{midblue}{rgb}{0,0.11372549,0.258823529}

\title{On the Limitation of Diffusion Models \\for Synthesizing Training Datasets}

\author{
Shin'ya Yamaguchi\textsuperscript{1,2}\thanks{Corresponding author. Email: \texttt{shinya.yamaguchi@ntt.com}}~~
Takuma Fukuda\textsuperscript{3}\thanks{Work done at NTT as an intern.}~~\\
\textsuperscript{1}NTT\quad\textsuperscript{2}Kyoto University\quad\textsuperscript{3}Chiba University
}

\begin{document}

\maketitle

\begin{abstract}
  Synthetic samples from diffusion models are promising for leveraging in training discriminative models as replications of real training datasets.
  However, we found that the synthetic datasets degrade classification performance over real datasets even when using state-of-the-art diffusion models.
  This means that modern diffusion models do not perfectly represent the data distribution for the purpose of replicating datasets for training discriminative tasks.
  This paper investigates the gap between synthetic and real samples by analyzing the synthetic samples reconstructed from real samples through the diffusion and reverse process.
  By varying the time steps starting the reverse process in the reconstruction, we can control the trade-off between the information in the original real data and the information added by diffusion models.
  Through assessing the reconstructed samples and trained models, we found that the synthetic data are concentrated in modes of the training data distribution as the reverse step increases, and thus, they are difficult to cover the outer edges of the distribution.
  Our findings imply that modern diffusion models are insufficient to replicate training data distribution perfectly, and there is room for the improvement of generative modeling in the replication of training datasets.
\end{abstract}

\begin{figure}[h]
    \centering
    \begin{minipage}{0.45\columnwidth}
        \includegraphics[width=\columnwidth]{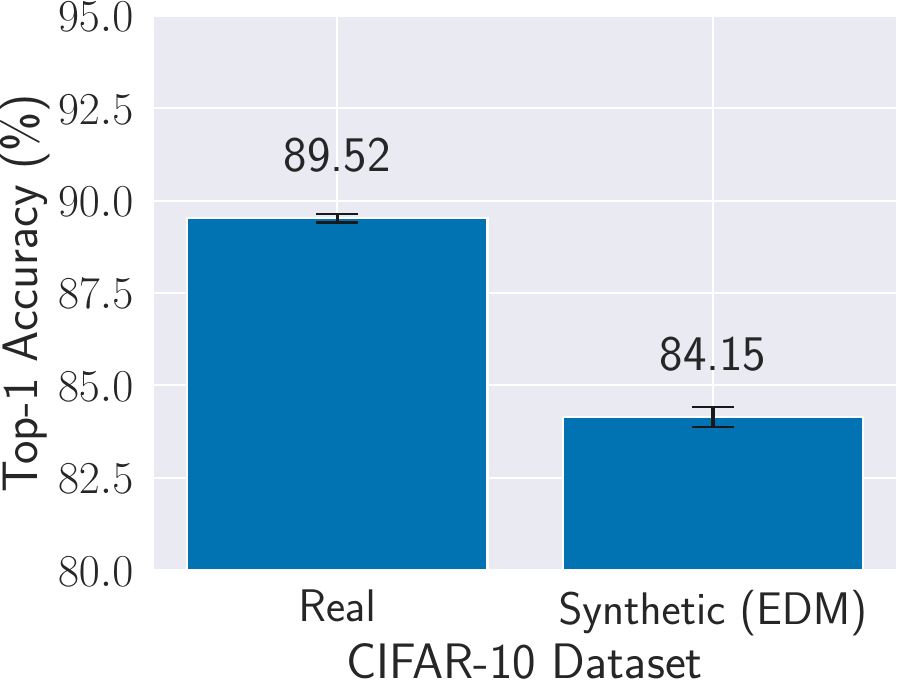}
        \subcaption{Top-1 Test Accuracy on CIFAR-10}\label{fig:top_performance}
    \end{minipage}
    \hfill
    \begin{minipage}{0.45\columnwidth}
        \includegraphics[width=\columnwidth]{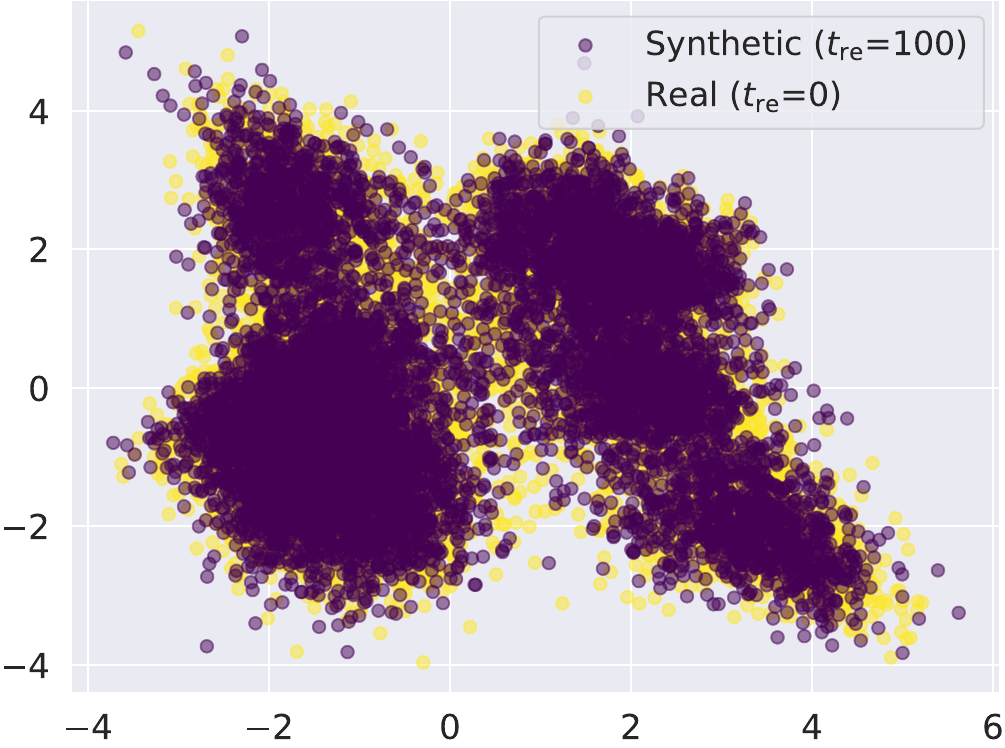}
    \subcaption{Feature Visualization (PCA) on ResNet-18}\label{fig:top_feature}
    \end{minipage}
    \caption{Our motivation and finding. (a): Synthetic datasets produced by a modern diffusion model (EDM~\cite{Karras_NeurIPS22_edm}) degrade the classification performance when solely using them for training classifiers. (b) We input synthetic samples to a classifier trained on a real dataset, and found that the features of synthetic samples concentrate on the modes of real feature distribution and do not cover the outer edge of the distribution. This means that diffusion models are still insufficient to replicate datasets for training classifiers.}
\end{figure}
\section{Introduction}\label{sec:intro}
In the past decade, deep generative models have witnessed remarkable advancements in generating high-quality synthetic samples that are human-indistinguishable from real data.
Among these generative models, diffusion models~\cite{ho_NeurIPS20_DDPM} have attracted much attention because they can outperform the existing generative models (e.g., GANs~\cite{Goodfellow_NIPS14_GANs} and VAEs~\cite{Kingma_ICLR14_VAE}) by learning reverse (denoising) processes through score-based likelihood maximization~\cite{dhariwal_NeurIPS21_diffusion_models_beat_gans,Rombach_CVPR22_StableDiffusion}.

The high-quality samples from diffusion models naturally raise research interest in their applicability for training target discriminative models (e.g., classifiers), and recent studies intensively develop training techniques utilizing synthetic samples from diffusion models.
For instance, He et al.~\cite{He_ICLR23_synthetic_zeroshot} demonstrated that synthetic samples from text-image diffusion models (e.g., Stable Diffusion~\cite{Rombach_CVPR22_StableDiffusion}) can achieve impressive zero-/few-shot learning performance by querying the synthetic training samples with crafted prompts representing target dataset categories.
Moreover, Burg et al.~\cite{Burg_arXiv23_diffusion_da_knn}, Azizi et al.~\cite{Azizi_arXiv23_diffusion_da_imagenet}, and Dunlap et al.~\cite{Dunlap_arXiv23_diffusion_da_text} highlighted the potential of diffusion models for data augmentation application. 
They investigated diffusion-based data augmentation methods by modifying diffusion models with nearest neighbor exploration in sampling~\cite{Brown_NeurIPS20_GPT3}, scaling up models~\cite{Azizi_arXiv23_diffusion_da_imagenet}, and customizing text prompts for querying samples~\cite{Dunlap_arXiv23_diffusion_da_text}.
However, in contrast to these remarkable successes, we observed that training models solely on na\"ively generated synthetic samples leads to inferior performance compared to models trained on real data (Figure~\ref{fig:top_performance}).
This indicates that even state-of-the-art diffusion models are not able to replicate the entire training data distribution, and there is a gap between real and synthetic datasets in terms of training classifiers.
In this paper, by analyzing diffusion models, we aim to answer the following important and open research question: \textit{What is the cause of the gap between real and synthetic datasets?}

This paper analyzes the gap between synthetic and real datasets from two perspectives: (i) the quality of synthetic samples and (ii) the impact of synthetic samples on training classification models.
To assess the gap, we introduce the concept of \textbf{real sample reconstruction} by focusing on the diffusion and reverse process.
Real sample reconstruction involves corrupting real samples by the diffusion process up to pre-defined steps and then restoring the corrupted samples by the reverse process; We refer to the pre-defined step as the reverse step.
By varying the reverse steps, we can continuously control the trade-off between the remaining information from the input real samples and the synthetic information injected by the reverse process (Figure~\ref{fig:reconstruct}).
We empirically investigate how the synthetic information affects the sample quality and the classification performance.

Our experimental findings are summarized as follows:
\begin{itemize}
    \item Diffusion models generate synthetic samples that are nearly indistinguishable as real or fake compared to competitive models such as GANs.
    \item Increasing the reverse steps (i.e., making sample properties closer to synthetic samples) leads to gradual degradation in the sample quality and classifier performance.
    \item Leveraging the synthetic samples for training classifiers does not adversely affect the tendency of classifier outputs (e.g., attention map).
    \item Synthetic samples are easier to classify than real samples.
    \item The synthetic samples concentrate near the modes of the data distribution in the feature space of classifiers (Fig.~\ref{fig:top_feature}), and a larger reverse step brings the sample closer to the mode.
\end{itemize}
These findings suggest that modern diffusion models have limitations in generating samples away from the modes.
This can be explained by the fact that diffusion models learn to denoise samples in the direction that maximizes the likelihood at each step in the reverse process.
That is, the reverse process may bring the sample closer to the typical mode.
Therefore, replicating the training dataset by diffusion models can result in accuracy degradation due to the less information far from the modes.


\section{Related Work}\label{sec:relatedwork}
Diffusion model~\cite{Sohl_ICML15_diffusion_origin, ho_NeurIPS20_DDPM} is a class of generative models inspired by thermodynamics.
They learn iteratively denoising process called \textit{reverse process} corresponding to the corruption process adding noises called \textit{diffusion process}.
Song et al.~\cite{Song_ICLR21_score} revealed the relationship between diffusion models and denoising score matching with stochastic gradient Langevin dynamics and explained optimization of the reverse process as score-based likelihood maximization.
By introducing conditional guidance in the reverse process, diffusion models successfully control output by class labels~\cite{dhariwal_NeurIPS21_diffusion_models_beat_gans, Ho_arXiv22_classifierfree} and text embedding~\cite{ramesh_2022_dalle2, Rombach_CVPR22_StableDiffusion}, and a number of subsequent studies are still being published. 

Since diffusion models can achieve high-quality synthetic samples in comparison to other generative models (e.g., GANs and VAEs)~\cite{dhariwal_NeurIPS21_diffusion_models_beat_gans}, recent studies investigated the capability of diffusion models as a source of training datasets~\cite{He_ICLR23_synthetic_zeroshot,Burg_arXiv23_diffusion_da_knn,Azizi_arXiv23_diffusion_da_imagenet,Dunlap_arXiv23_diffusion_da_text}.
These studies utilized text-image pre-trained diffusion models such as Stable Diffusion~\cite{Rombach_CVPR22_StableDiffusion} for generating synthetic training samples and succeeded in improving classification performance by adding the synthetic samples into training datasets.
In contrast, this paper focuses on class conditional diffusion models trained only on target datasets for discriminative tasks and does not consider large pre-trained diffusion models to discard the effects of knowledge transfer from external pre-trained datasets.

\section{Preliminary}\label{sec:preliminary}
Here, we briefly introduce the principles of diffusion models and real sample reconstruction, which is used for our main analysis.
\subsection{Diffusion Models}\label{sec:diffusion_model}
A diffusion model learns a data distribution \(p(x)\) by optimizing the parameterized reverse (denoising) process assuming Markov Chain with length \(T\)~\cite{Sohl_ICML15_diffusion_origin,ho_NeurIPS20_DDPM}, which corresponds to the forward diffusion process.
Specifically, most modern diffusion models are optimized by minimizing the family of the following loss function with respect to the neural network parameter \(\theta\)~\cite{ho_NeurIPS20_DDPM,dhariwal_NeurIPS21_diffusion_models_beat_gans,Rombach_CVPR22_StableDiffusion}.
\begin{equation}\label{eq:loss_diffusion}
\mathcal{L}(\theta) = \mathbb{E}_{x,\epsilon\sim\mathcal{N}(0,1), t} \left[ \|\epsilon - \epsilon_\theta(x_t, t)\|^2_2 \right],
\end{equation}
where \(\epsilon_\theta\) is the denoising autoencoder parameterized by \(\theta\), \(t\) is the time step randomly sampled from \(\{1,\cdots,T\}\), \(x\) is the input, and \(x_t\) is a noisy variant of \(x\).
In inference time, a synthetic sample \(\hat{x}\) is generated by sequentially applying the denoising function  for each \(t\) from \(T\) to \(1\) as 
\begin{equation}
    x_{t-1} = \frac{1}{\sqrt{\alpha_t}}\left(x_t-\frac{1-\alpha_t}{\sqrt{1-\bar{\alpha}_t}}\epsilon_\theta(x_t,t) + \sigma_t z\right),
\end{equation}
where \(\alpha_t = 1 - \beta_t\), \(\beta_t\) is a scheduled variance in \(\{\beta_1,\cdots,\beta_T\}\), \(\bar{\alpha}_t = \prod^t_{s=1}\alpha_s\), \(\sigma_t = \sqrt{\frac{1-\bar{\alpha}_{t-1}}{1-\bar{\alpha}_t}\beta_t}\) and \(z\sim\mathcal{N}(0,1)\).
Song et al.~\cite{Song_ICLR21_score} showed that this denoising process corresponds to stochastic gradient Langevin dynamics, which produces samples by iterative updating \(x_t\) with the score \(\nabla_x\log p(x)\):
\begin{equation}\label{eq:stochastic_Langevin}
    x_{t} = x_{t-1} + \frac{\delta}{2}\nabla_x\log p(x_{t-1}) + \sqrt{\delta} z,
\end{equation}
where \(\delta\) is a step size.
In this paper, we implement diffusion models with conditional EDM~\cite{Karras_NeurIPS22_edm} to generate a synthetic labeled dataset for training classifiers.

\begin{figure}[t]
    \centering
    \begin{minipage}{0.5\columnwidth}
    \begin{algorithm}[H]
        \caption{Real Sample Reconstruction}\label{alg:real_sample_reconstruction}
        \begin{algorithmic}[1]
        {\small
            \REQUIRE{Real sample \(x\), reverse step \(t_\mathrm{re} > 1\)}
            \ENSURE{Reconstructed sample \(\hat{x}\)}
            \STATE{// Corrupting \(x\) with diffusion process for \(t_\mathrm{re}\)}
            \STATE{\(x_0 \leftarrow x\)}
            \FOR{\(t=1,\cdots,t_\mathrm{re}\)}
            \STATE \(x_t \leftarrow \sqrt{{\alpha}_t} x_{t-1} + \sqrt{1-\alpha_t}\epsilon_{t-1}\)
            \ENDFOR
            \STATE{// Restoring \(x_{t_\mathrm{re}}\) with reverse process}
            \FOR{\(t=t_\mathrm{re},\cdots,1\)}
            \STATE \(\hat{x}_{t-1} \leftarrow \frac{1}{\sqrt{\alpha_t}}\left(x_t-\frac{1-\alpha_t}{\sqrt{1-\bar{\alpha}_t}}\epsilon_\theta(x_t,t) + \sigma_t z\right)\)
            \ENDFOR
            \STATE \(\hat{x}\leftarrow \hat{x}_0\) 
        }
        \end{algorithmic}
    \end{algorithm}
    \end{minipage}
    \hfill
    \begin{minipage}{0.45\columnwidth}
        \includegraphics[width=\columnwidth]{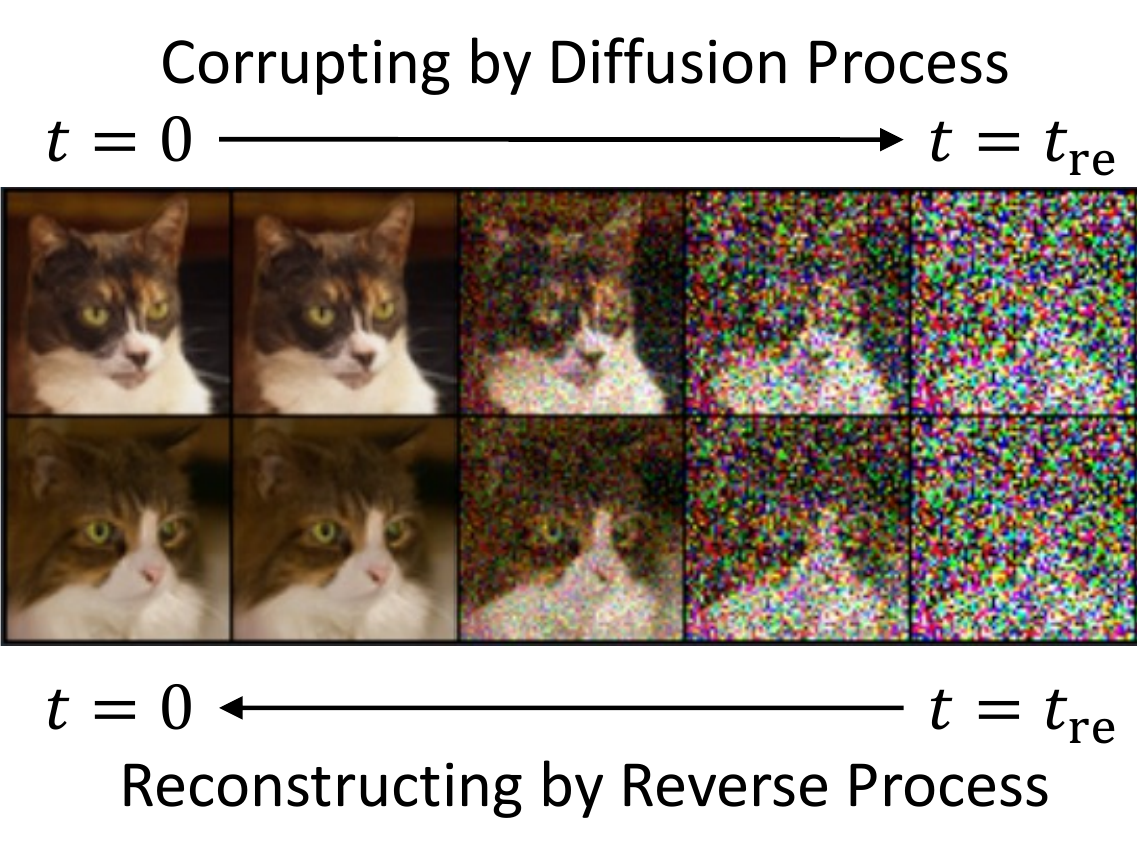}
    \caption{Real Sample Reconstruction}\label{fig:reconstruct}
    \end{minipage}
\end{figure}

\subsection{Real Sample Reconstruction}
We introduce real sample reconstruction, which produces intermediate samples between real and synthetic by exploiting the diffusion and reverse process.
Real sample reconstruction first corrupts the real samples by the diffusion process from \(0\) to a specified time step \(t_\mathrm{re}\), and then recovers the corrupted samples by the reverse process from \(t_\mathrm{re}\) to \(0\).
Given a real data point \(x\), we produce a reconstructed sample \(\hat{x}\) with a reverse time step \(t_\mathrm{re}\) by following Algorithm~\ref{alg:real_sample_reconstruction}.
This reconstruction algorithm is similar to SDEdit~\cite{Meng_ICLR22_sdedit}, which is an image-editing method based on the reconstruction of guide images by diffusion models.
Intuitively, \(\hat{x}\) is fully real when \(t_\mathrm{re}=0\), a fully synthetic when \(t_\mathrm{re}=T\), and half of real and synthetic when \(t_\mathrm{re}=T/2\) as depicted in Figure~\ref{fig:reconstruct}.
Unlike the purpose of SDEdit, we aim to produce intermediate samples of real and synthetic by simply inputting real images into the diffusion and reverse process.

\section{Analysis}
In this section, we report the experimental results assessing (i) the quality of reconstructed samples from diffusion models and (ii) the effects on classifiers trained on the reconstructed samples.
We used the CIFAR-10 dataset~\cite{krizhevsky09_cifar10} as the target dataset, the CIFAR-10 pre-trained conditional EDM~\cite{Karras_NeurIPS22_edm} (\(T=100\)) as the diffusion model, and ResNet-18~\cite{he_resnet} as the classifier architecture.

\subsection{Analysis on Synthetic Sample}
\paragraph{Evaluation Protocol.}
To analyze reconstructed synthetic samples, we measured Frech\`et inception distance (FID)~\cite{heusel_ttur_fid_nips17}, precision/recall ~\cite{kynkaanniemi_NeurIPS19_improved_precision_and_recall}, and fake detection accuracy~\cite{frankICML20_leveraging_frequency}.
Among them, FID and precision/recall are measured on the ImageNet pre-trained feature extractor.
FID evaluates the gap between real and synthetic datasets, and precision/recall evaluate the probabilities that synthetic/real samples fall within the real/synthetic distributions.
Fake detection accuracy is calculated on a classifier trained to distinguish real and synthetic samples on both the pixel and frequency domains.
This is useful to find out how different synthetic and real samples are in terms of input to the classifier.
We used 50,000 synthetic samples and 50,000 real samples to calculate the metrics.

\begin{figure}[t]
    \centering
    \begin{minipage}{0.45\columnwidth}
        \centering
        \includegraphics[width=0.9\columnwidth]{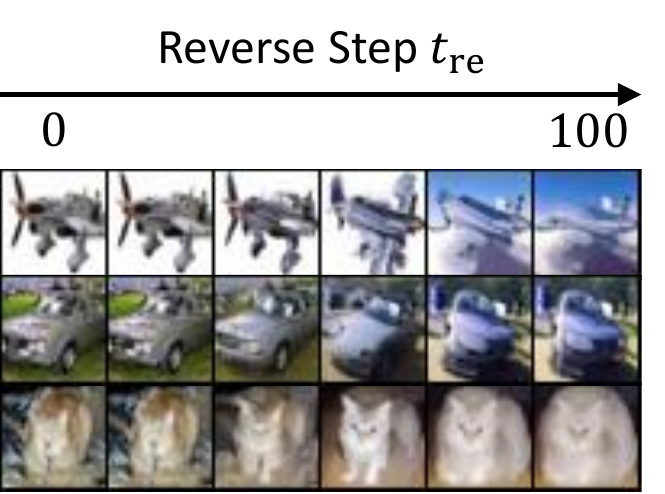}
        \subcaption{Reconstructed Samples}\label{fig:reconstructed_samples}
    \end{minipage}
    \hfill
    \begin{minipage}{0.45\columnwidth}
        \centering
        \includegraphics[width=\columnwidth]{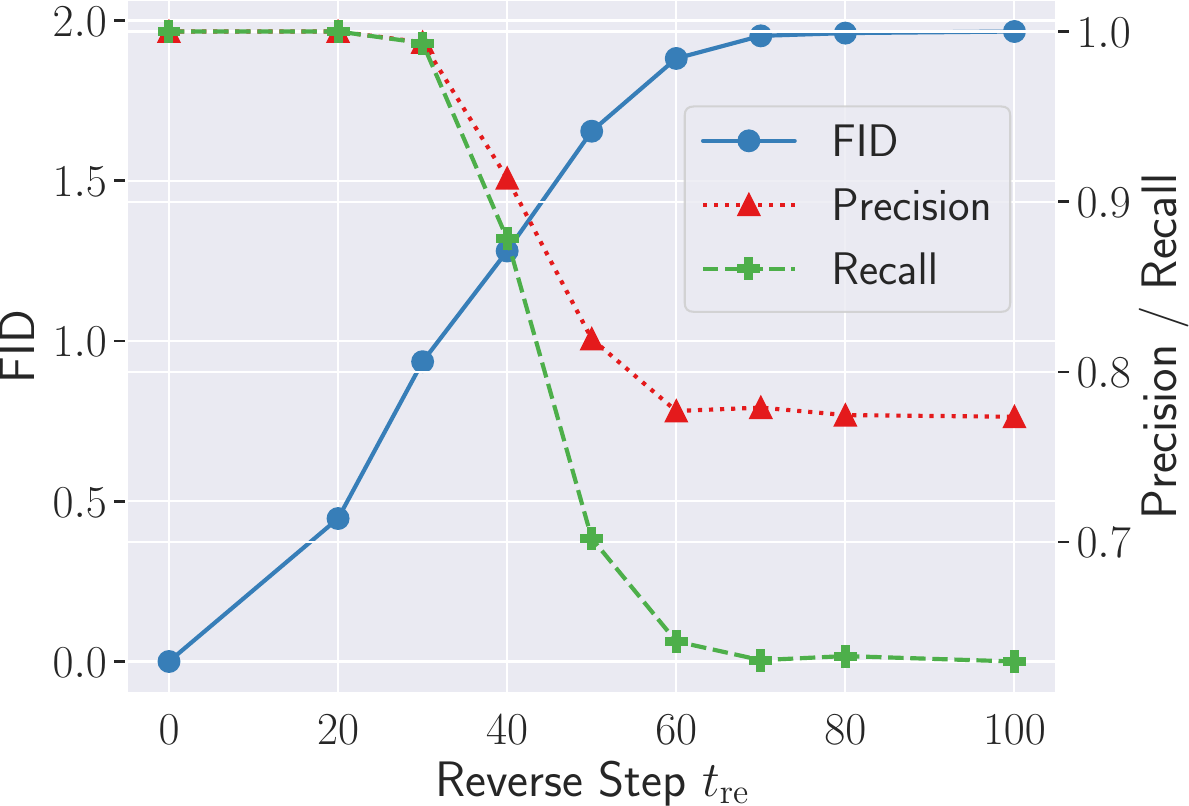}
        \subcaption{FID~\cite{heusel_ttur_fid_nips17} and Precision/Recall~\cite{kynkaanniemi_NeurIPS19_improved_precision_and_recall}}\label{fig:fid_pcrc}
    \end{minipage}
    \caption{Quality Assessments of Synthetic Sample}
    \vspace{-3mm}
\end{figure}

\paragraph{Sample Quality.}
We first show the visualization of the reconstructed samples in Figure~\ref{fig:reconstructed_samples}.
We reconstructed the samples on every 10 steps of \(t_\mathrm{re}\in [20,80]\)  from EDM.
As the reverse step \(t_\mathrm{re}\) increases, the reconstructed samples gradually lose information on the input real sample, and represent information on the synthetic sample.
Nevertheless, in visual quality, it is hard to distinguish between a synthetic sample and a real sample for every \(t_\mathrm{re}\).
Next, we show the FID and precision/recall scores calculated on the reconstructed samples in Figure~\ref{fig:fid_pcrc}.
We see that increasing reverse steps progressively degrades all the quantitative metrics.
This indicates that the reverse process may be harmful to maintain the information on the real samples.
In particular, the recall scores are significantly degraded by the reverse process, indicating that the synthetic sample does not sufficiently cover the training data distribution.

\begin{wraptable}{r}{0.3\columnwidth}
    \vspace{-3mm}
    \centering
    \caption{Fake Detection Accuracy (CIFAR-10)}
    \label{tab:fake_detection_acc}
            \resizebox{0.3\columnwidth}{!}{
        \begin{tabular}{lcc}\toprule
            \multirow{2}{*}{Generative Model} & \multicolumn{2}{c}{Accuracy (\%)} \\
             & Pixel & DCT \\
          \midrule
            StyleGAN3~\cite{Karras_NeurIPS21_stylegan3} & 89.56 & 53.62\\
            EDM~\cite{Karras_NeurIPS22_edm} & 56.15 & 58.91\\
            \bottomrule
        \end{tabular}}
\end{wraptable}
\paragraph{Fake Detection Accuracy.} We demonstrate the fake detection accuracy on the synthetic samples.
To evaluate the worst quality case, we used \(t_\mathrm{re}=100\) in this experiment.
Table~\ref{tab:fake_detection_acc} shows the fake detection accuracy in the pixel domain and frequency domain (DCT).
For comparison, we also print the result of StyleGAN3~\cite{Karras_NeurIPS21_stylegan3}.
The higher scores mean easier samples to be detected as fake.
While the StyleGAN3 samples were easily distinguished by the classifier, fewer EDM samples were detected as fake.
These results suggest that although the synthetic datasets from diffusion models differ in quantitative measures such as FID, their properties as input to the classifier are almost the same as those of the real samples.

\begin{figure}[h]
    \centering
    \begin{minipage}{0.45\columnwidth}
        \centering
        \captionof{table}{Top-1 Test Accuracy on CIFAR-10}
        \label{tab:top1acc}
        \begin{tabular}{lcc}\toprule
        Reverse Step \(t_\mathrm{re}\) & Test Acc. (\%) \\\midrule
        0 (Fully Real) & 89.52\(^{\pm\text{.11}}\)\\
        30 & 88.84\(^{\pm\text{.08}}\)\\
        50 & 87.41\(^{\pm\text{.32}}\)\\
        70 & 85.76\(^{\pm\text{.49}}\)\\
        100 (Fully Synthetic) & 84.15\(^{\pm\text{.27}}\) \\\bottomrule
        \end{tabular}
    \end{minipage}
    \hfill
    \begin{minipage}{0.45\columnwidth}
        \centering
        \includegraphics[width=\columnwidth]{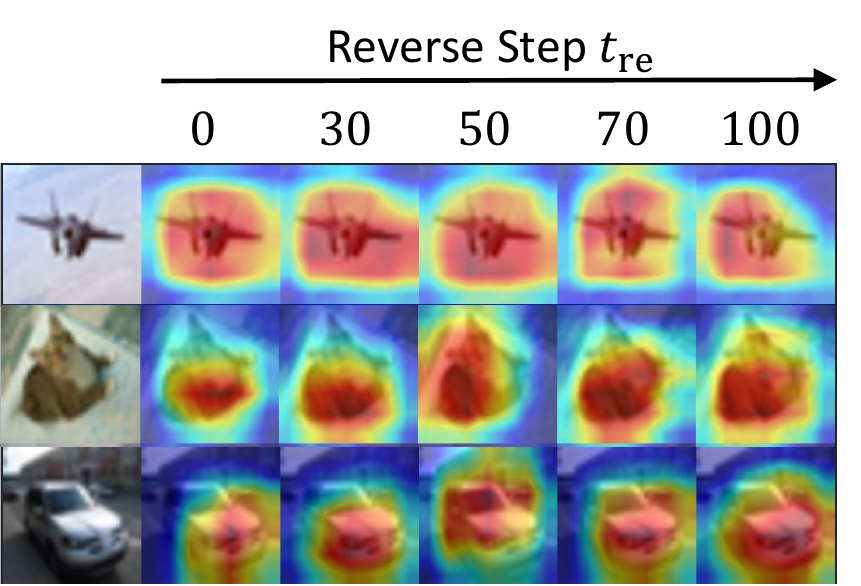}
        \caption{GradCAM Visualization}\label{fig:gradcam}
    \end{minipage}
    \vspace{-3mm}
\end{figure}

\subsection{Analysis on Training Classifiers}
\paragraph{Evaluation Protocol.}
We analyze trained classifiers on reconstructed synthetic samples by varying the reverse step \(t_\mathrm{re}\).
We evaluate test classification accuracy, attention map visualization by GradCAM~\cite{Selvaraju_ICCV17_GradCAM}, output entropy, and feature visualization by principle component analysis (PCA).
We trained ResNet-18 classifiers for 100 epochs on the synthetic CIFAR-10 datasets yielded by real sample reconstruction with \(t_\mathrm{re} = 30, 50, 70\), and tested them on the real CIFAR-10 test set.
We used the SGD optimizer with a learning rate of \(0.01\) dropping by multiplying \(0.1\) for every 30 epochs.
We also show the results when using the real dataset (i.e., \(t_\mathrm{re}=0\)) and the fully synthetic dataset by the reverse process with random noise (i.e., \(t_\mathrm{re}=100\)).
For GradCAM and feature visualization, we used the output of \texttt{block4} on ResNet-18.
We calculate the marginal output entropy by
\begin{equation}
     H_\theta(y) = - \frac 1 N \sum_i^N \sum_j^{C}  p_\theta(y = j | x_i) \log (p_\theta(y = j | x_i)),
\end{equation}
where \(N\) is a dataset size, \(C\) is a class number, \(p_\theta (y=j | x_i) = \frac{\exp(f_\theta(x_i)[j])}{\sum_k^C \exp(f_\theta(x_i)[k])}\), \(f_\theta\) is a classifier.\looseness-1

\paragraph{Classification Performance.}
Table~\ref{tab:top1acc} shows the top-1 test accuracy on the real CIFAR-10 test set for each reverse time step \(t_\mathrm{re}\).
Similar to the sample quality shown in the previous section, we see that the performance of the classifier degrades as the reverse step increases.
This implies that the reverse process of diffusion models eliminates information important for solving classification tasks from the original real sample.

\paragraph{Attention Map.}
Figure~\ref{fig:gradcam} shows the visualizations of GradCAM.
We input real test samples of CIFAR-10 for each trained model.
Interestingly, while the test accuracy is degraded by real sample reconstruction, the synthetic samples used for training do not change the attention of the trained models.
This means that the synthetic sample itself has no noticeable negative impact on learning classification tasks.

\begin{figure}[t]
    \centering
    \begin{minipage}{0.45\columnwidth}
        \centering
        \includegraphics[width=\columnwidth]{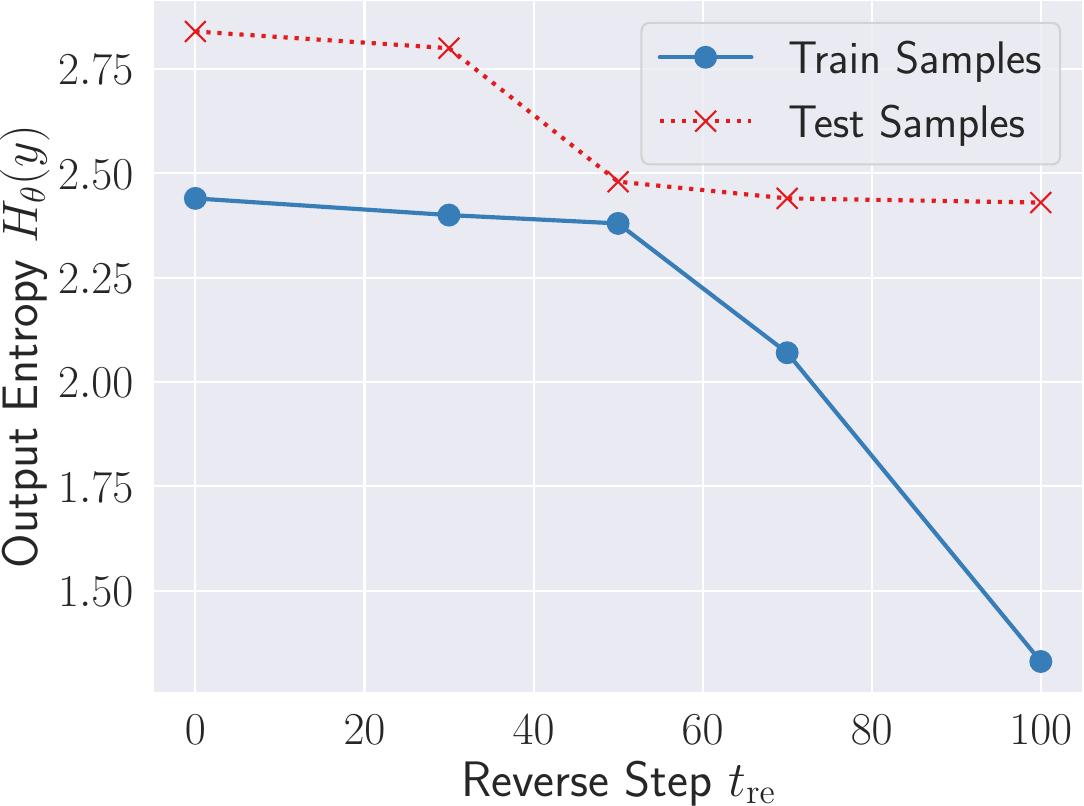}
        \caption{Output Entropy of Classifier}\label{fig:entropy}
    \end{minipage}
    \hfill
    \begin{minipage}{0.45\columnwidth}
        \centering
        \includegraphics[width=\columnwidth]{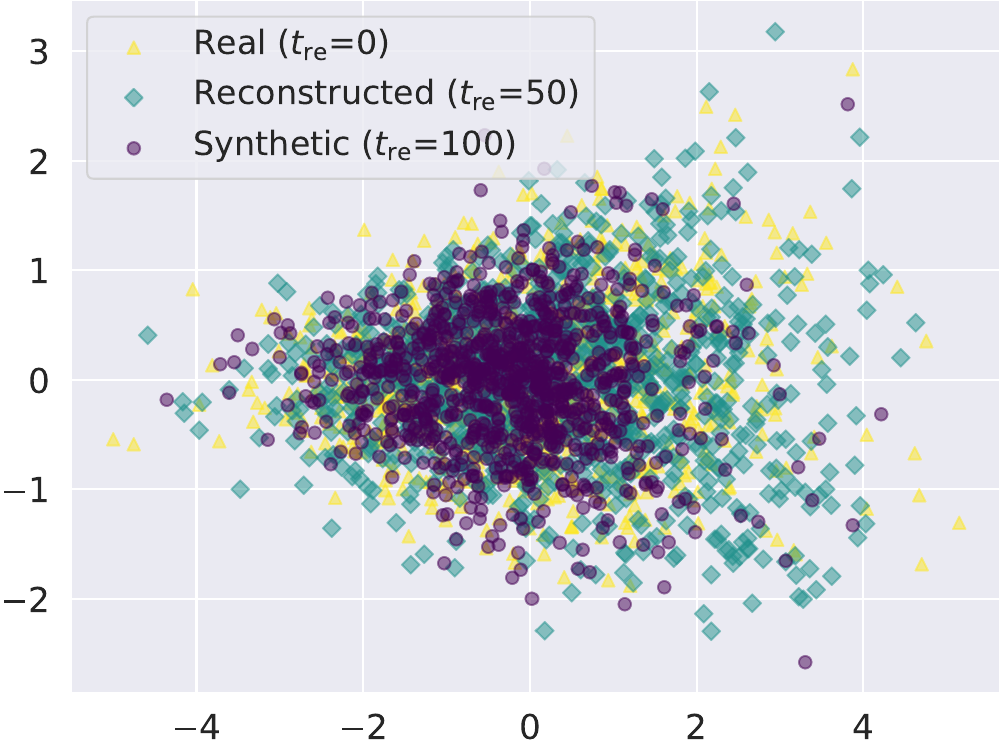}
        \caption{Feature Visualization (\texttt{truck} class)}\label{fig:pca_feat}
    \end{minipage}
    \vspace{-3mm}
\end{figure}

\paragraph{Output Entropy}
Next, we assess the quantitative effects on the classifier's prediction.
To this end, we used the classifier trained on the real CIFAR-10 because we can consider it an ideal classifier for the purpose of training dataset replication.
Figure~\ref{fig:entropy} plots the output entropy \(H_\theta(y)\).
We calculated \(H_\theta(y)\) by inputting the synthetic samples into the classifier.
In Figure~\ref{fig:entropy}, \texttt{Train Samples} and \texttt{Test Samples} mean the calculated entropy scores on the reconstructed samples from real samples of the train/test set.
Note that again, in this experiment, we used only the classifier trained on the real CIFAR-10 to assess the characteristics of synthetic samples.
We see that increasing the reverse step makes the synthetic samples low entropy, indicating easy samples to be classified.
Thus, the diffusion model tends to produce a typical sample that is representative of the class by the reverse process.

\paragraph{Feature Visualization.}
Finally, we visualize the features of synthetic samples to examine how synthetic samples behave on the classifier.
Similar to the previous paragraph, we used the classifier trained on the real CIFAR-10 for feature visualization.
We applied PCA to the extracted features of input synthetic samples and reduced the dimension to two.
Figure~\ref{fig:top_feature}~and~\ref{fig:pca_feat} are the visualization results of all class samples and \texttt{truck} class samples, respectively.
In Figure~\ref{fig:top_feature}, the synthetic sample is concentrated inside the distribution formed by the real samples, while its outer edges are not well covered.
Meanwhile, in Figure~\ref{fig:pca_feat}, the reconstructed at \(t_\mathrm{re}=50\) appears to cover the region where the sample at \(t_\mathrm{re}=100\) is scarce.
These results suggest that the synthetic samples from diffusion models tend to concentrate the center (mode) of training data distribution, and the reverse process gradually pulls the synthetic samples toward the modes of training distribution.

\section{Discussion}
Through the empirical analysis in the previous section, we observed that the modern diffusion models can produce quite realistic synthetic samples, but they still have insufficient generative performance for replicating training datasets for classifiers.
In particular, the reverse process of diffusion models seems to gradually concentrate the synthetic samples toward the modes of the training data distribution.
This can be explained by the interpretation of the diffusion model as a score-based generative model.
As we discussed in Sec.~\ref{sec:diffusion_model}, a reverse process corresponds to a step of stochastic gradient Langevin dynamics as shown in Eq.~\ref{eq:stochastic_Langevin}.
That is, a reverse step contains the gradient of log-likelihood (score) \(\nabla_x\log p(x)\).
Therefore, the iterative denoising of samples by multiple reverse steps means that the samples are moving closer to the region of greater likelihood, i.e., the mode of the distribution.
Eq. (3) has a disturbance term by $z$ to prevent the concentration of sampling near the modes, but our experimental results suggest that this cannot be completely prevented for the purpose of replicating training datasets.

\section{Conclusion and Takeaway}\label{sec:conclusion}
This paper empirically showed the limitations of diffusion models for synthesizing datasets for training classifiers.
Modern diffusion models are not sufficient to replicate entire training datasets due to the sampling concentration near the data distribution modes.
This can be caused by the reverse denoising process, which naturally moves the samples toward the modes.
From these observations, one of the important takeaways is that we should improve diffusion models to cover the outside edges of training data distributions. 
Another one is that, currently, the data augmentation applications of diffusion models, which utilize both real and synthetic samples, can be more suitable to train high-performance classifiers than replicating entire training datasets and utilizing only synthetic samples.
We believe that these observations and implications will be helpful for future research.

\clearpage
\bibliography{ref}

\begin{thebibliography}{10}

\bibitem{Karras_NeurIPS22_edm}
Tero Karras, Miika Aittala, Timo Aila, and Samuli Laine.
\newblock Elucidating the design space of diffusion-based generative models.
\newblock In {\em Advances in neural information processing systems}, 2022.

\bibitem{ho_NeurIPS20_DDPM}
Jonathan Ho, Ajay Jain, and Pieter Abbeel.
\newblock Denoising diffusion probabilistic models.
\newblock In {\em Advances in neural information processing systems}, 2020.

\bibitem{Goodfellow_NIPS14_GANs}
Ian Goodfellow, Jean Pouget-Abadie, Mehdi Mirza, Bing Xu, David Warde-Farley, Sherjil Ozair, Aaron Courville, and Yoshua Bengio.
\newblock Generative adversarial nets.
\newblock In {\em Advances in Neural Information Processing Systems 27}, 2014.

\bibitem{Kingma_ICLR14_VAE}
Diederik~P. Kingma and Max Welling.
\newblock Auto-encoding variational bayes.
\newblock In {\em International Conference on Learning Representations}, 2014.

\bibitem{dhariwal_NeurIPS21_diffusion_models_beat_gans}
Prafulla Dhariwal and Alex Nichol.
\newblock Diffusion models beat gans on image synthesis.
\newblock In {\em Advances in Neural Information Processing Systems}, 2021.

\bibitem{Rombach_CVPR22_StableDiffusion}
Robin Rombach, Andreas Blattmann, Dominik Lorenz, Patrick Esser, and Bj\"orn Ommer.
\newblock High-resolution image synthesis with latent diffusion models.
\newblock In {\em Proceedings of the IEEE/CVF Conference on Computer Vision and Pattern Recognition}, 2022.

\bibitem{He_ICLR23_synthetic_zeroshot}
Ruifei He, Shuyang Sun, Xin Yu, Chuhui Xue, Wenqing Zhang, Philip Torr, Song Bai, and Xiaojuan Qi.
\newblock Is synthetic data from generative models ready for image recognition?
\newblock {\em arXiv preprint arXiv:2210.07574}, 2022.

\bibitem{Burg_arXiv23_diffusion_da_knn}
Max~F Burg, Florian Wenzel, Dominik Zietlow, Max Horn, Osama Makansi, Francesco Locatello, and Chris Russell.
\newblock A data augmentation perspective on diffusion models and retrieval.
\newblock {\em arXiv preprint arXiv:2304.10253}, 2023.

\bibitem{Azizi_arXiv23_diffusion_da_imagenet}
Shekoofeh Azizi, Simon Kornblith, Chitwan Saharia, Mohammad Norouzi, and David~J Fleet.
\newblock Synthetic data from diffusion models improves imagenet classification.
\newblock {\em arXiv preprint arXiv:2304.08466}, 2023.

\bibitem{Dunlap_arXiv23_diffusion_da_text}
Lisa Dunlap, Alyssa Umino, Han Zhang, Jiezhi Yang, Joseph~E Gonzalez, and Trevor Darrell.
\newblock Diversify your vision datasets with automatic diffusion-based augmentation.
\newblock {\em arXiv preprint arXiv:2305.16289}, 2023.

\bibitem{Brown_NeurIPS20_GPT3}
Tom Brown, Benjamin Mann, Nick Ryder, Melanie Subbiah, Jared~D Kaplan, Prafulla Dhariwal, Arvind Neelakantan, Pranav Shyam, Girish Sastry, Amanda Askell, et~al.
\newblock Language models are few-shot learners.
\newblock {\em Advances in neural information processing systems}, 2020.

\bibitem{Sohl_ICML15_diffusion_origin}
Jascha Sohl-Dickstein, Eric Weiss, Niru Maheswaranathan, and Surya Ganguli.
\newblock Deep unsupervised learning using nonequilibrium thermodynamics.
\newblock In {\em International conference on machine learning}, 2015.

\bibitem{Song_ICLR21_score}
Yang Song, Jascha Sohl-Dickstein, Diederik~P Kingma, Abhishek Kumar, Stefano Ermon, and Ben Poole.
\newblock Score-based generative modeling through stochastic differential equations.
\newblock In {\em International Conference on Learning Representation}, 2021.

\bibitem{Ho_arXiv22_classifierfree}
Jonathan Ho and Tim Salimans.
\newblock Classifier-free diffusion guidance.
\newblock {\em arXiv preprint arXiv:2207.12598}, 2022.

\bibitem{ramesh_2022_dalle2}
Aditya Ramesh, Prafulla Dhariwal, Alex Nichol, Casey Chu, and Mark Chen.
\newblock Hierarchical text-conditional image generation with clip latents.
\newblock {\em arXiv preprint arXiv:2204.06125}, 2022.

\bibitem{Meng_ICLR22_sdedit}
Chenlin Meng, Yutong He, Yang Song, Jiaming Song, Jiajun Wu, Jun-Yan Zhu, and Stefano Ermon.
\newblock Sdedit: Guided image synthesis and editing with stochastic differential equations.
\newblock In {\em International Conference on Learning Representation}, 2022.

\bibitem{krizhevsky09_cifar10}
Alex Krizhevsky and Geoffrey Hinton.
\newblock Learning multiple layers of features from tiny images.
\newblock Technical report, Citeseer, 2009.

\bibitem{he_resnet}
Kaiming He, Xiangyu Zhang, Shaoqing Ren, and Jian Sun.
\newblock Deep residual learning for image recognition.
\newblock In {\em Proceedings of the IEEE conference on computer vision and pattern recognition}, 2016.

\bibitem{heusel_ttur_fid_nips17}
Martin Heusel, Hubert Ramsauer, Thomas Unterthiner, Bernhard Nessler, and Sepp Hochreiter.
\newblock Gans trained by a two time-scale update rule converge to a local nash equilibrium.
\newblock In {\em Advances in Neural Information Processing Systems}, 2017.

\bibitem{kynkaanniemi_NeurIPS19_improved_precision_and_recall}
Tuomas Kynk{\"a}{\"a}nniemi, Tero Karras, Samuli Laine, Jaakko Lehtinen, and Timo Aila.
\newblock Improved precision and recall metric for assessing generative models.
\newblock {\em Advances in Neural Information Processing Systems}, 2019.

\bibitem{frankICML20_leveraging_frequency}
Joel Frank, Thorsten Eisenhofer, Lea Sch{\"o}nherr, Asja Fischer, Dorothea Kolossa, and Thorsten Holz.
\newblock Leveraging frequency analysis for deep fake image recognition.
\newblock 2020.

\bibitem{Karras_NeurIPS21_stylegan3}
Tero Karras, Miika Aittala, Samuli Laine, Erik H{\"a}rk{\"o}nen, Janne Hellsten, Jaakko Lehtinen, and Timo Aila.
\newblock Alias-free generative adversarial networks.
\newblock {\em Advances in Neural Information Processing Systems}, 2021.

\bibitem{Selvaraju_ICCV17_GradCAM}
Ramprasaath~R Selvaraju, Michael Cogswell, Abhishek Das, Ramakrishna Vedantam, Devi Parikh, and Dhruv Batra.
\newblock Grad-cam: Visual explanations from deep networks via gradient-based localization.
\newblock In {\em Proceedings of the IEEE international conference on computer vision}, 2017.

\end{thebibliography}
\bibliographystyle{unsrt}

\end{document}